\documentclass[10pt,twocolumn,letterpaper]{article}

\usepackage[pagenumbers]{cvpr} 

\definecolor{cvprblue}{rgb}{0.21,0.49,0.74}
\usepackage[pagebackref,breaklinks,colorlinks,allcolors=cvprblue]{hyperref}
\usepackage{amssymb}
\usepackage{booktabs}
\usepackage{multirow}

\usepackage{bbding}           
\usepackage{marvosym}         

\usepackage{textcomp}
\usepackage{tablefootnote}
\usepackage[accsupp]{axessibility}
\usepackage{graphicx}
\usepackage{xcolor}
\usepackage{float}
\usepackage{colortbl}

\usepackage[most]{tcolorbox}
\usepackage{enumitem}

\newtcolorbox{promptbox}[1]{
  enhanced,
  breakable,
  colback=white,
  colframe=black,
  colbacktitle=black,
  coltitle=white,
  fonttitle=\bfseries,
  title=#1,
  boxrule=0.8pt,
  left=1.5mm,
  right=1.5mm,
  top=1.5mm,
  bottom=1.5mm,
}

\hypersetup{pagebackref,breaklinks,colorlinks}

\def\paperID{10321} 
\def\confName{CVPR}
\def\confYear{2026}

\title{4DWorldBench: A Comprehensive Evaluation Framework for 3D/4D World Generation Models}

\author{
Yiting Lu\textsuperscript{1,*}, Wei Luo\textsuperscript{1,*}, Peiyan Tu\textsuperscript{2,3,*}, Haoran Li\textsuperscript{1,*}, Hanxin Zhu\textsuperscript{1,3,*}, Zihao Yu\textsuperscript{1}, \\
Xingrui Wang\textsuperscript{1}, Xinyi Chen\textsuperscript{1}, Xinge Peng\textsuperscript{1}, Xin Li\textsuperscript{1}, Zhibo Chen\textsuperscript{1,3,\dag} \\ [2mm]
\textsuperscript{1} University of Science and Technology of China \\
\textsuperscript{2} Zhejiang University \quad
\textsuperscript{3} Beijing Zhongguancun Academy \\
{\tt \small \{luyt31415, lw21, lihr, hanxinzhu\}@mail.ustc.edu.cn, pytu@zju.edu.cn} \\
{\tt \small \{xin.li, chenzhibo\}@ustc.edu.cn}
}
\begin{document}
\maketitle
\let\thefootnote\relax\footnotetext{* Equal contribution \quad \dag Corresponding author}

\begin{abstract}
World Generation Models are emerging as a cornerstone of next-generation multimodal intelligence systems. Unlike traditional 2D visual generation, World Models aim to construct realistic, dynamic, and physically consistent 3D/4D worlds from images, videos, or text. These models not only need to produce high-fidelity visual content but also maintain coherence across space, time, physics, and instruction control, enabling applications in virtual reality, autonomous driving, embodied intelligence, and content creation.
However, prior benchmarks emphasize different evaluation dimensions and lack a unified assessment of world-realism capability.
To systematically evaluate World Models, we introduce the 4DWorldBench, which measures models across four key dimensions: Perceptual Quality, Condition–4D Alignment, Physical Realism, and 4D Consistency. 
The benchmark covers tasks such as Image-to-3D/4D, Video-to-4D, Text-to-3D/4D. Beyond these, we innovatively introduce adaptive conditioning across multiple modalities, which not only integrates but also extends traditional evaluation paradigms. To accommodate different modality-conditioned inputs, we map all modality conditions into a unified textual space during evaluation, and further integrate LLM-as-judge, MLLM-as-judge, and traditional network-based methods. This unified and adaptive design enables more comprehensive and consistent evaluation of alignment, physical realism, and cross-modal coherence.
Preliminary human studies further demonstrate that our adaptive tool selection achieves closer agreement with subjective human judgments.
We hope this benchmark will serve as a foundation for objective comparisons and improvements, accelerating the transition from "visual generation" to "world generation." Our project can be found at https://yeppp27.github.io/4DWorldBench.github.io/.


\end{abstract}

\vspace{-3mm}

\section{Introduction}
\label{sec:intro}

\textbf{World Generation Models} are rapidly emerging as a new paradigm for multimodal intelligence. In contrast to conventional visual generation, which typically optimizes frame fidelity or short-horizon temporal coherence, World Models aim to construct \emph{realistic, dynamic, and physically consistent 3D/4D worlds} from text, image, or video conditions. Beyond high perceptual quality, World Models are expected to maintain \emph{spatial--temporal coherence}, \emph{physical realism}, and \emph{interactive controllability}, enabling applications across \emph{virtual reality}, \emph{autonomous driving}, \emph{embodied AI}.

Existing benchmarks for generative models can be broadly categorized into three categories: video-centric, world generation-focused, and physics-aware. Video-based benchmarks (e.g., VBench2.0~\cite{vbench2025}) emphasize high-level properties such as fidelity, creativity, commonsense, physics and controllability by leveraging broad prompt suites and hierarchical-level evaluations. However, they rely heavily on manually predefined question templates and offer only coarse evaluation of physical realism and 4D consistency. World generation benchmarks (e.g., WorldScore~\cite{duan2025worldscore}) go beyond  video generation to evaluate controllable 3D/4D generation across modalities, offering structured scene specifications and standardized metrics. While effective in assessing 3D/4D scene generation ability, they still exhibit limitations in fine-grained condition-4D alignment and provide insufficient coverage of physical realism. Lastly, physics-aware benchmarks (e.g., PhyGenBench~\cite{meng2024phygenbench}) explicitly focus on testing adherence to physical laws through curated prompts and rule-based evaluation pipelines. Although valuable for physical reasoning, these benchmarks often lack modality diversity, semantic richness, and scalability due to handcrafted logic. These gaps underscore the need for a more comprehensive, adaptive, and semantically grounded benchmark—such as our proposed 4DWorldBench—capable of evaluating 3D/4D world generation across diverse modalities and dimensions.

In short, the community lacks a \emph{general-purpose}, \emph{multimodal}, and \emph{physics-aware} benchmark that measures \emph{perceptual quality}, \emph{condition--4D alignment}, \emph{physical realism}, and \emph{4D consistency} in a single, extensible framework. In this work, we propose a more unified benchmark for world generation, named \textbf{4DWorldBench}. Our goal is to provide a comprehensive evaluation of \emph{world generation models} that go beyond traditional visual synthesis, targeting the generation of coherent \textbf{3D and 4D worlds}. Specifically, 4DWorldBench supports models conditioned on multiple modalities, including \textbf{text}, \textbf{image}, and  \textbf{video}, thereby enabling fair comparison across diverse input settings. Moreover, recognizing that world generation inherently involves not only perceptual fidelity but also compliance with the laws of nature, we explicitly introduce evaluation along the \textbf{physical realism dimension}, in addition to perceptual quality, condition--4D alignment, and 4D consistency. By combining these perspectives, 4DWorldBench aims to establish a more complete and challenging standard for assessing the next generation of world generation systems.

To support fine-grained evaluation across key dimensions—particularly condition–4D alignment and physical realism—we combine two complementary evaluation tracks. First, we employ an Multimodal Large Language Models (MLLM)-based Question Answer (QA) evaluator that inspects the generated video and answers a set of dimension-specific questions derived from the input condition (converted into text). 
Second, to handle more abstract reasoning—especially for assessing high-level physical realism—we introduce an LLM-based QA evaluator. 
This dual-track design is motivated by the observation that current MLLMs, while proficient at surface-level video question answering, often struggle with complex physical reasoning (e.g., fluid dynamics, force interactions). In contrast, LLMs excel at abstract reasoning and compositional understanding when provided with high-quality textual descriptions. By leveraging the strengths of both model types, our framework enables a more robust and semantically aware evaluation across a broad spectrum of world generation scenarios. Moreover, some dimensions need hybrid methods for evaluation: feature-based similarity metrics can be easily misled by videos containing only subtle object motions, resulting in inflated alignment scores; therefore, we additionally incorporate an MLLM-based QA to guard against such deceptive cases.
Our work offers three contributions:
\begin{itemize}
    \item \textbf{Comprehensive, multimodal benchmark for world generation.} 4DWordBench evaluates four main dimensions and multiple sub-dimensions, with balanced text/image/video condition sets for both non-physical and physical categories.
    \item \textbf{Adaptive hybrid evaluation methodology.} We integrate traditional metrics with LLM/MLLM-driven question decomposition and answering, enabling fine-grained, condition-aware assessment of alignment and physics while improving scalability.
    \item \textbf{Unified, extensible framework and leaderboard.} 4DWorldBench supports Image-to-3D/4D, Video-to-4D, Text-to-3D/4D, providing transparent, reproducible comparisons across models.
\end{itemize}

\begin{table*}[t]
\centering
\caption{Evaluation dimensions in our 4DWorldBench. Four main dimensions expand into multiple sub-dimensions.}
\label{tab:dimensions}
\begin{tabular}{l | c}
\toprule
\textbf{Main Dimension} & \textbf{Sub-dimensions} \\
\midrule
Perceptual Quality & Spatial Quality; Temporal Quality; 3D Texture \\
Condition-4D Alignment & Event Control; Scene Control; Attribute Relationship Control; Motion Control \\
Physical Realism & Dynamics; Optics; Thermal \\
4D Consistency & Viewpoint Consistency; Motion Consistency; Style Consistency \\
\bottomrule
\end{tabular}
\vspace{-3mm}
\end{table*}

\begin{table}[t]
\centering
\caption{Comparison with representative benchmarks. ``Modalities'' refer to supported \emph{condition} types. ``Eval.\ dims'' mark whether the benchmark covers Perceptual Quality (Q), Condition Alignment (A), Physical Realism (P), and 4D Consistency (C). ``\checkmark" indicates full support for the dimension, ``partial'' denotes limited or partially covered evaluation, and ``--'' indicates that the dimension is not supported.
}
\label{tab:comparison}
\resizebox{0.9\linewidth}{!}{\begin{tabular}{l c c c c c}
\toprule
\textbf{Benchmark} & \textbf{Condition Modalities} & \textbf{Q} & \textbf{A} & \textbf{P} & \textbf{C} \\
\midrule
WorldScore~\cite{duan2025worldscore} & Text & \checkmark & \textit{partrial} & -- & \checkmark \\
VBench~\cite{huang2024vbench} & Text & \checkmark & \checkmark & \textit{partrial} & --  \\
VBench2.0~\cite{vbench2025} & Text / Image & \checkmark & \checkmark & \textit{partrial} & --  \\
PhyGenBench~\cite{meng2024phygenbench} & Text & -- & -- & \checkmark & -- \\
\textbf{4DWorldBench (ours)} & \textbf{Text / Image / Video} & \checkmark & \checkmark & \checkmark & \checkmark \\

\bottomrule
\end{tabular}}
\vspace{-5mm}
\end{table}

%
\vspace{-1mm}
\section{Related Work}
\label{sec:related}
\vspace{-1mm}
\subsection{Video-Based Evaluation Benchmarks}
\vspace{-1mm}
A large body of work focuses on evaluating video generation models in terms of fidelity, coherence, and semantic correctness~\cite{li2025worldmodelbench,gu2025phyworldbench,vbench2025}. 
VBench~\cite{huang2024vbench} introduced a comprehensive benchmark that decomposes video quality into fine-grained dimensions such as identity consistency, motion smoothness, and temporal flicker, with scores aligned to human preferences. 
Its successor, VBench-2.0~\cite{vbench2025}, extends coverage to five key aspects: human fidelity, controllability, creativity, physics, and commonsense, leveraging vision--language models for deeper evaluation. 
Other works target specific challenges: DEVIL~\cite{liao2024evaluation} evaluates visual vividness and the honesty of video content under dynamic settings; and T2V-CompBench~\cite{compbench2024} and TC-Bench~\cite{tcbench2024} examine compositionality in multi-object, multi-attribute videos.  
Perceptual quality has also been tackled via large-scale frameworks such as EvalCrafter~\cite{evalcrafter2023}, which integrates objective metrics and human ratings into a unified leaderboard. Other benchmarks also highlight narrative or scene-level evaluation: StoryEval~\cite{storyeval2024} measures story completion in multi-event prompts using VLM-based evaluators, showing that models struggle beyond a few coherent events.
These efforts provide complementary views, but each focuses on a subset of evaluation dimensions.
\vspace{-1mm}
\subsection{3D/4D World Generation Benchmarks}
\vspace{-1mm}
Evaluating 3D/4D generation requires capturing long-horizon spatial and temporal consistency. 
Benchmarks for text-to-3D scene generation: GT23D-Bench~\cite{su2024gt23d}) 
introduces a large-scale, well-organized 3D dataset with multimodal annotations and comprehensive 3D-aware metrics for evaluating text–3D alignment and visual quality, but are limited to static snapshots. 
And Eval3D~\cite{duggal2025eval3d} provides a fine-grained and interpretable evaluation framework that assesses 3D generation quality across semantic and geometric dimensions using consistency among diverse foundation models and tools.
WorldScore~\cite{duan2025worldscore} frames the task as sequential world generation with explicit instructions (e.g., camera trajectories), evaluating controllability, quality, and dynamics across multi-scene sequences. 
Overall, these benchmarks focus on measuring whether models can generate persistent, controllable, and temporally consistent 3D/4D worlds.
\vspace{-2mm}
\subsection{Physics-Focused Video Benchmarks}
\vspace{-1mm}
Recent advancements in video generation have led to the development of several benchmarks aimed at evaluating models' understanding of physical dynamics. PhyGenBench~\cite{meng2024phygenbench} designs diservse prompts across multiple fundamental physical laws, including mechanics, optics, material science, and thermal properties. It provides a PhyGenEval pipeline that combines vision-language models and rule-based QA to assess physical plausibility. VideoPhy~\cite{videophy2024} and VideoPhy-2~\cite{bansal2025videophy} construct the dataset with explicit physical rules, testing models' ability to understand physical interactions. PisaBench~\cite{li2025pisa} focuses on real-world free-fall tasks to evaluate the ability of generative models to reproduce physical phenomena like gravity and dynamic collisions. LLMPhy \cite{cherian2024llmphy} integrates large language models with physics engines for advanced physical reasoning, using datasets for predicting stable poses in complex object interactions.  Physics-IQ~\cite{motamed2025generative} evaluates generative video models by giving them the initial frames of a real physical scene, having them predict the next few seconds, and then comparing the generated video against ground-truth recordings across multiple physics-based metrics.


\begin{figure*}[tbp]
  \centering
   \includegraphics[width=1.0\linewidth]{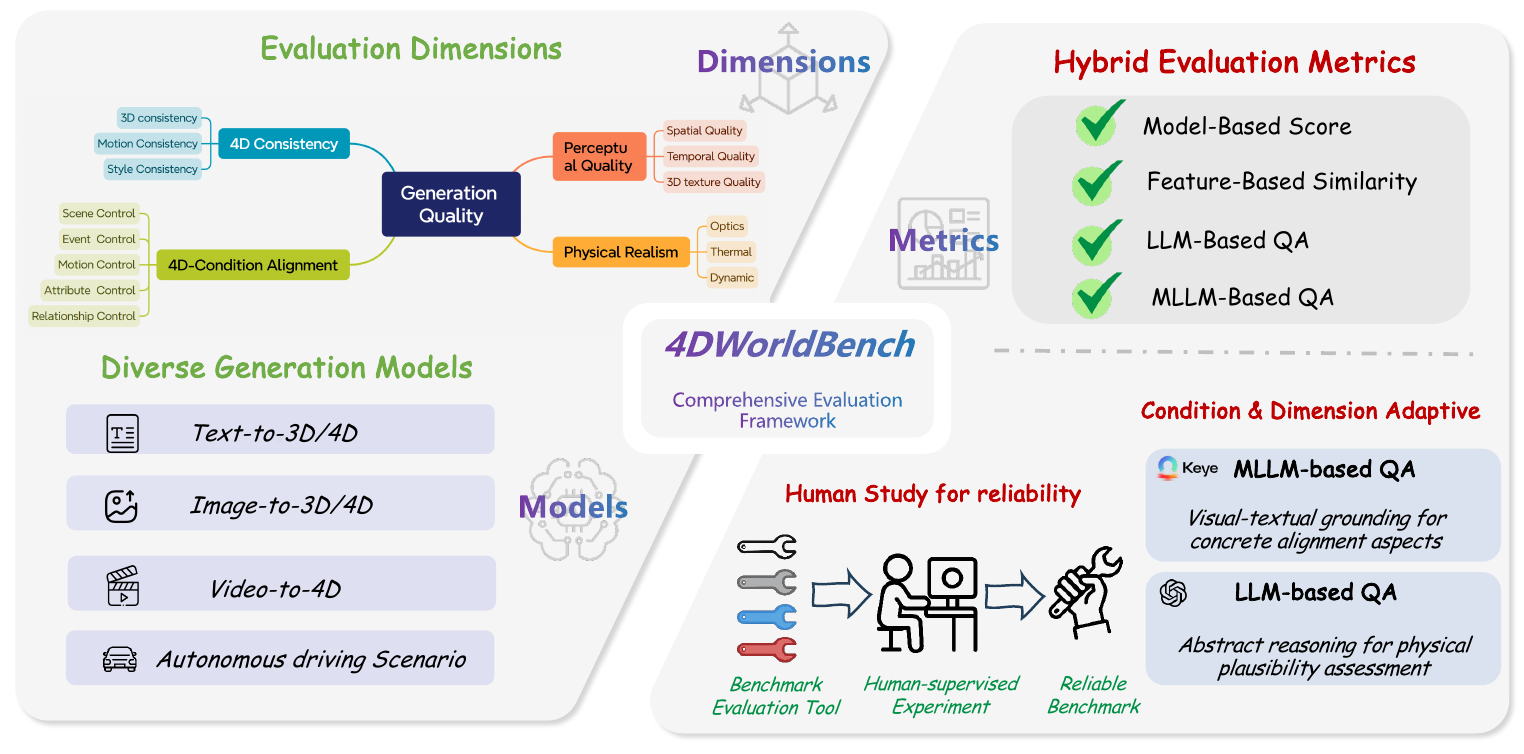}
   \vspace{-6mm} 
   \caption{Overview of the 4DWorldBench framework. The benchmark evaluates generation quality across four key dimensions: 4D consistency, condition alignment, perceptual quality, and physical realism. It supports diverse generative settings, including text-, image-, and video-to-3D/4D generation. The framework integrates hybrid evaluation metrics—model-based scores, feature-based similarity, LLM-based QA, and MLLM-based QA—together with human studies for reliability. Condition- and dimension-adaptive QA modules leverage MLLMs for concrete visual grounding and LLMs for higher-level physical reasoning.
}
   \vspace{-3mm} 
   \label{fig:demo_T2I}
\end{figure*}

\section{Benchmark Dataset}


To ensure comprehensive and controllable evaluation, we curate a benchmark dataset that spans both physics and non-physics domains, across text, image, and video modalities. The data collection process emphasizes category diversity, and physical plausibility.

\begin{figure}[tbp]
  \centering
   \includegraphics[width=0.95\linewidth]{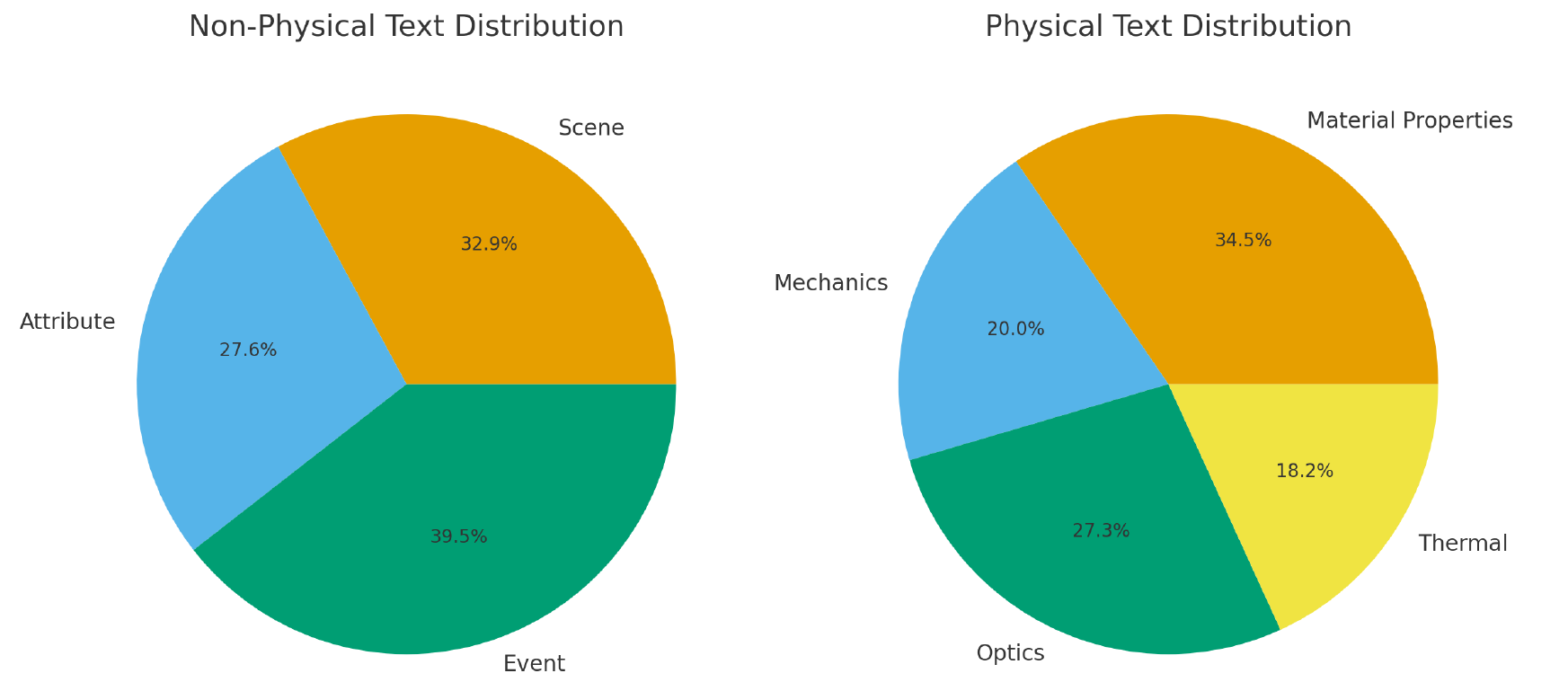}
   \vspace{-3mm} 
   \caption{Text Condition Distribution.}
   \vspace{-3mm} 
   \label{fig:con_text}
\end{figure}

\noindent \textbf{Text Condition Collection.} For non-physical modality, text prompts are sampled from WorldScore~\cite{duan2025worldscore}, a benchmark that decomposes world generation into next-scene prediction tasks and explicitly distinguishes between static and dynamic scenes. We sample the text condition across several categories based on the structure and semantics of the text prompts: 
\textbf{Scene, Event, and Attribute}, which is shown in Fig.~\ref{fig:con_text}  
For physical modality, to evaluate physical reasoning and control, we incorporate physics-driven prompts and exemplars following a structured taxonomy (see Fig.~\ref{fig:con_text}), which includes four major classes: Material, Mechanics, Optics, and Thermal phenomena. Each class is further divided into representative subclasses—for example, Optics includes refraction, reflection, and the Tyndall effect; Mechanics includes gravity, buoyancy, and pressure; Material covers color, hardness, combustibility, and redox reactions; and Thermal involves heating and phase transitions. These Text prompts are revised from PhyGenBench~\cite{meng2024phygenbench}. We further utilize GPT-5 to produce data augmentations for thermal physics.
In total, the evaluation set includes 50 physical text conditions and 76 non-physical conditions in text form.
\begin{figure}[tbp]
  \centering
   \includegraphics[width=0.95\linewidth]{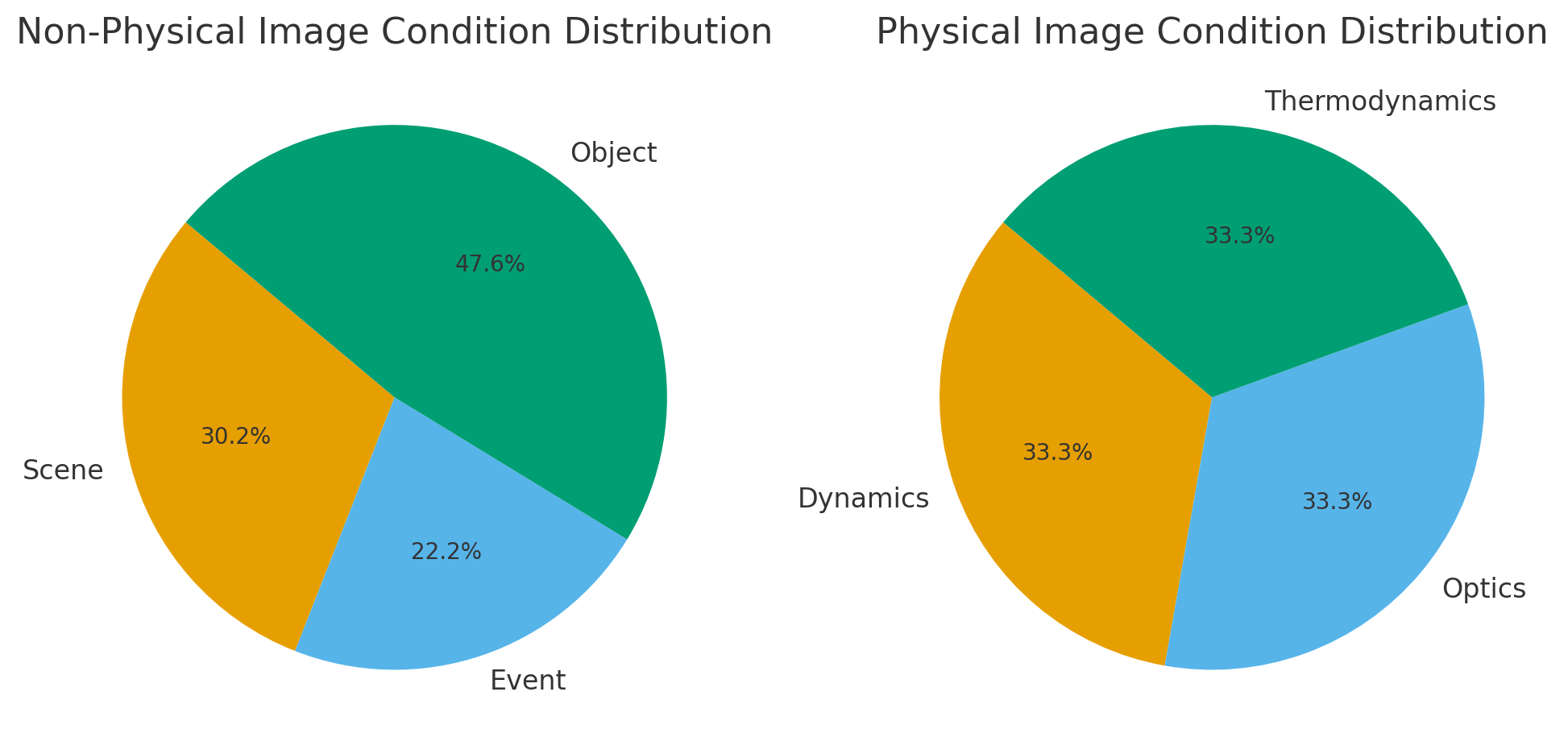}
   \vspace{-5mm} 
   \caption{Image Condition Distribution.}
   \vspace{-4mm} 
   \label{fig:con_image}
\end{figure}
\noindent \textbf{Image Condition Collection.}
For non-physical condition, we sampled image conditions from Vbench2.0~\cite{vbench2025}. We sample the image conditions from Vbench2.0 across several categories based on the structure and semantics of the captions: 
\textbf{Scene, Event, Object}. 
Specifically, captions involving interactive entities or motion are classified as \textbf{Event}, those describing a single interacting entity are labeled as \textbf{Object}, and captions without explicit entities are assigned to the \textbf{Scene} category.
For object-level 3D generation model only, we sample 25 object-centric images (without background) from Objaverse-XL dataset~\cite{deitke2023objaverse}. For physical condition, We first collect videos from \textit{WISA}~\cite{wang2025wisa}, after which the first frame of each selected video is extracted to serve as the corresponding image conditioning input. The distribution of physical and non-physical image conditioning samples is shown in Fig.~\ref{fig:con_image}. Among them, 48 samples correspond to physical conditions, while 63 belong to non-physical scene conditions.
\begin{figure}[tbp]
  \centering
   \includegraphics[width=0.95\linewidth]{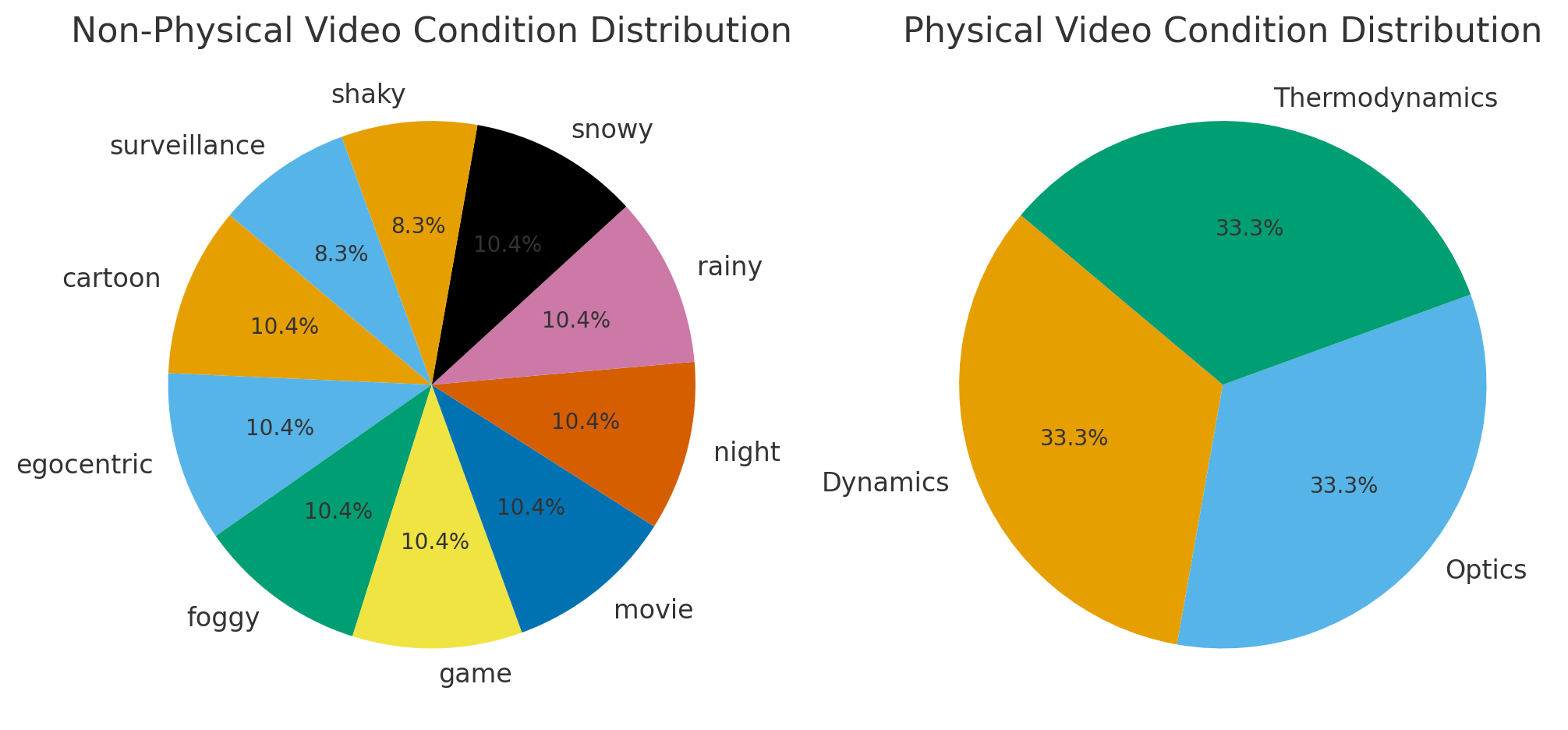}
   \vspace{-5mm} 
   \caption{Video Condition Distribution.}
   \vspace{-8mm} 
   \label{fig:con_video}
\end{figure}
\noindent \textbf{Video Condition Collection.}
For non-physical condition, we source captioned video clips from WideRange4D~\cite{yang2025widerange4d}. These datasets provide diverse annotations, including scene type, weather condition, and crowd density. Our sampling strategy proceeds in two stages. First, we ensure coverage across stylistic and domain categories—such as cartoon, game, night, foggy, egocentric, shaky, and surveillance views. Second, within each category, we further stratify the samples by temporal motion amplitude, measured using inter-frame optical flow statistics. Accordingly, the videos are categorized into \textit{low-motion} and \textit{high-motion} types. This two-stage sampling process promotes domain diversity while ensuring comprehensive coverage of motion control capabilities.

For physical condition, we collect videos from \textit{WISA}~\cite{wang2025wisa}, which covers three physical domains—\textbf{Dynamic}, \textbf{Optic}, and \textbf{Thermodynamic}.
We first uniformly sample videos from these three categories to ensure balanced domain coverage.
Within each category, we further stratify the data into \textit{high-motion} and \textit{low-motion} subsets based on temporal motion amplitude.
The distribution of physical and non-physical video conditioning samples is shown in Fig.~\ref{fig:con_video}. Each subset contains 48 samples respectively.


\begin{figure*}[tbp]
  \centering
   \includegraphics[width=0.9\linewidth]{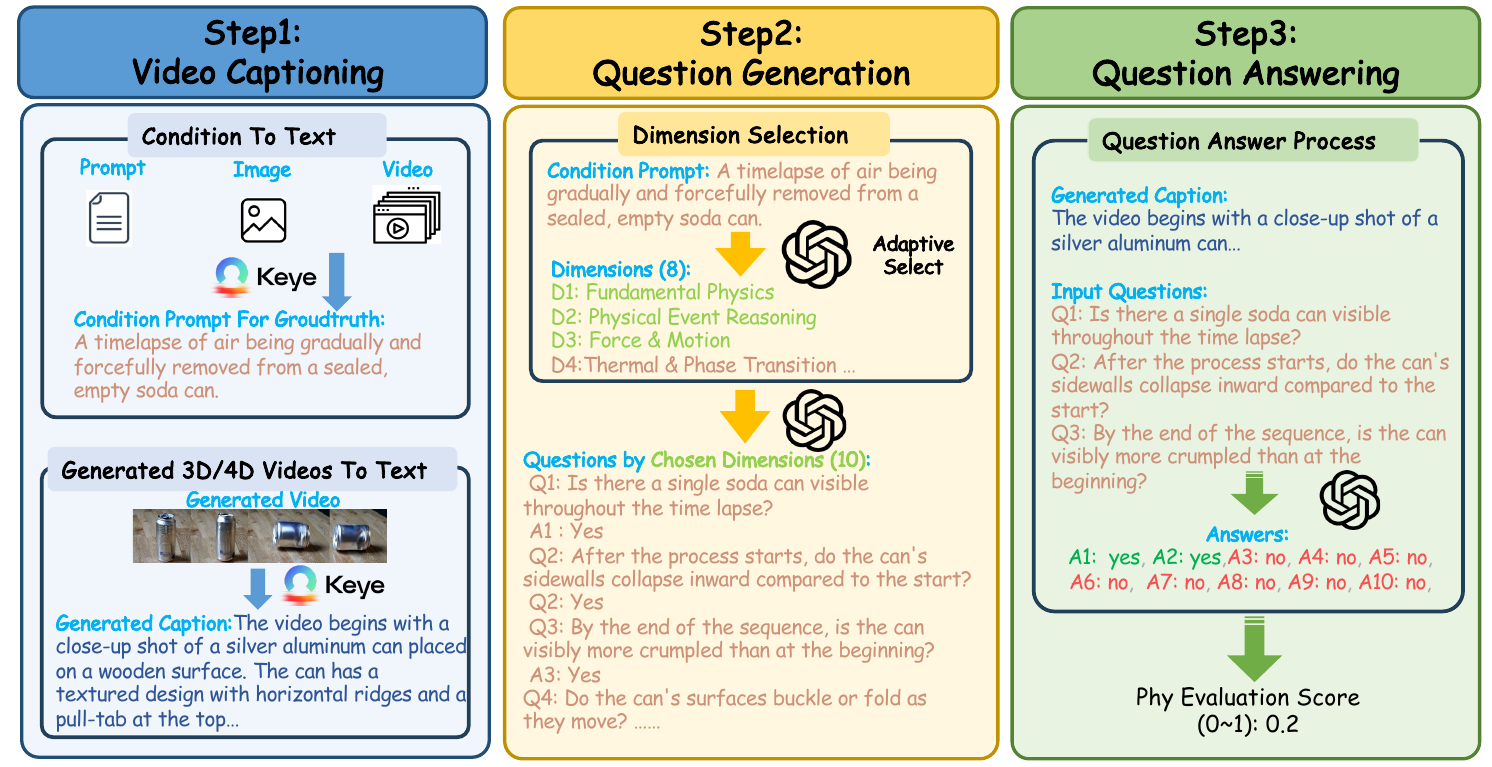}
   \vspace{-3mm} 
   \caption{Pipeline of the Physical Realism Evaluation. The framework first unifies text, image, and video conditions into a common textual form, then uses an LLM to adaptively select relevant physical dimensions and generate diagnostic questions. An LLM-based QA module compares predicted and reference answers to yield a continuous physical realism score.}
   \vspace{-3mm} 
   \label{fig:phy}
\end{figure*}

\begin{table*}[htbp]
\centering
\scriptsize
\caption{Leaderboard for different categories of 4D generation models.}
\vspace{-3mm}
\resizebox{0.98\linewidth}{!}{\begin{tabular}{l|ccc|ccccc|ccc|ccc|c}
\toprule
\multirow{2}{*}{\textbf{Model}} 
& \multicolumn{3}{c|}{\textbf{Perceptual Quality}} 
& \multicolumn{5}{c|}{\textbf{Condition-4D Alignment}} 
& \multicolumn{3}{c|}{\textbf{Physical Realism}} 
& \multicolumn{3}{c|}{\textbf{4D Consistency}} 
& \multirow{2}{*}{\textbf{Overall}} \\ 
\cmidrule(lr){2-4} \cmidrule(lr){5-9} \cmidrule(lr){10-12} \cmidrule(lr){13-15}
& Spatial & Temporal & 3D Texture 
& Event & Scene & Attribute & relationship & Motion 
& Dynamics & Optics & Thermal 
& Viewpoint & Motion & Style 
& \\ 
\midrule

\multicolumn{16}{l}{\textbf{Image-to-4D}} \\ 
CamI2V~\cite{zheng2024cami2v}             & 0.419+0.524& 0.668&0.700 &0.397 &0.954  & 0.922 & 0.845& 0.756&0.650&0.738&0.629&0.701 & 0.553+0.816& 0.887&  0.697
\\ 
DiffusionAsShader~\cite{gu2025diffusion}  &0.509+0.522 &0.790 &0.715 &0.400 &0.965 & 0.963 &0.804& 0.781&0.688&0.850&0.763&0.908 &  0.874+0.750 & 0.918&  0.763
\\ 

\midrule
\multicolumn{16}{l}{\textbf{Video-to-4D}} \\ 
EX-4D~\cite{ex4d}         &0.305+0.409 &0.356 & 0.616 &0.615 &0.805 &0.792&0.879 & 0.869 &0.421&0.627&0.600& 0.266&  0.384+0.811 &0.827 &  0.599
\\ 
ReCamMaster~\cite{bai2025recammaster}    &0.215+0.401 &0.394 &0.636 & 0.573& 0.757 &0.673&0.811 & 0.796 &0.680&0.714&0.773& 0.862&0.859+0.834 &0.985 & 0.685
\\
TrajectoryCrafter~\cite{yu2025trajectorycrafter} &0.492+0.434 &0.560 & 0.646&0.639 &0.862 &0.766&0.913 &0.856 &0.671&0.667&0.636&  0.362&0.454+0.883&0.878 &  0.670
\\ 
Vista~\cite{gao2024vista}        &0.216+0.402& 0.397&0.657 &0.405&0.686 &0.516&0.698& 0.607 &0.607&0.613&0.593& 0.921&0.942+0.622 &  0.986&  0.617
\\ 

\midrule
\multicolumn{16}{l}{\textbf{Text-to-4D}} \\ 
4Dfy~\cite{4dfy}           & 0.336+0.469 &0.442 & 0.569& 0.566& 0.505 & 0.326&0.553& 0.545& 0.265&0.417&0.200& 0.741 & 0.934+0.705&0.993&  0.535
\\ 
dreamin4D~\cite{dreamin4D}         & 0.234+0.396 &0.181 & 0.655 & 0.409&0.576 & 0.390&0.558& 0.416&0.235&0.238&0.150&0.612 & 0.761+0.360&0.850 &  0.439\\ 


\bottomrule
\end{tabular}}
\vspace{-1mm}
\end{table*}

\section{Benchmark Metric}

Our proposed 4DWorldBench measures 3D/4D generation models across four key dimensions: Physical Realism, Condition-4D Alignment, and 4D Consistency, Perceptual Quality.

\subsection{Physical Realism}
Our physical realism evaluation framework is built upon two core principles: (1) converting both the input conditions (except for text conditions) and the generated videos into captions, leveraging LLMs to assess physical consistency through caption-based reasoning instead of direct video observation; and (2) adaptively selecting evaluation dimensions based on the semantics of each input caption to ensure scenario-relevant physical assessment.
Specially, to evaluate the degree of physical realism in generated videos, we design a three-step framework that combines video captioning and condition captioning, physics-aware question generation, and question answering for quantitative scoring. 

\textbf{1) Video Captioning for Condition and Generated Video.} 
The input condition $C$, originating from any modality other than text (e.g., image, video), is first transformed into a textual description of the underlying physical process through Keye-VL 1.5~\cite{yang2025kwaikeyevl}. The generated 4D video $V$ is subsequently captioned using Keye-VL 1.5 to obtain $\hat{T}$, providing a unified textual representation for downstream physical reasoning.
\textbf{2) Question Answer Generation.} In the second step, we construct a set of diagnostic questions $\{Q_i\}_{i=1}^N$ derived from both the condition prompt $C$ and canonical physical laws. These questions are designed to cover different dimensions of physical understanding, including fundamental physics, physical event reasoning, force and motion, and thermal or phase transitions. For example, one question might ask whether the sidewalls of a soda can collapse inward once air is removed. Such questions target observable outcomes that must hold if the generated video is physically plausible.
 \textbf{3) Scoring through QA.} In the third step, we input the generated caption $\hat{T}$ along with the question set $\{Q_i\}$ into a large language model, which produces predicted answers $\{\hat{A}_i\}$. These predictions are then compared with the expected physical outcomes $\{A_i^\ast\}$. For each question, we assign a binary correctness score defined as $ s_i = \mathbb{1}(\hat{A}_i = A_i^\ast)$.
And the overall physical realism score is computed as the mean accuracy across all questions: $
S_{\text{phy}} = \frac{1}{N} \sum_{i=1}^N s_i, \quad S_{\text{phy}} \in [0,1]$.
A higher value of $S_{\text{phy}}$ indicates that the generated video adheres more closely to physical reality, while a lower score reflects violations of basic physical constraints, such as the absence of expected deformations, incorrect motion dynamics, or conservation law inconsistencies. More details about prompt and questions can be seen in \textbf{supplementary}.

\subsection{4D Condition Alignment}

The Condition Alignment evaluation framework measures how well a generated video conforms to its specified condition, ensuring consistency with the intended storyline (event), character motion (motion), attribute and scene. It follows a systematic three-step process— condition input captioning, question generation, and question answering—to quantitatively evaluate alignment quality.
\textbf{1) Condition Captioning.}
The input condition is first converted into a textual description.
\textbf{2) Question Generation.}
We then construct a set of diagnostic questions probing key alignment aspects, including object/character attributes, spatial relations, scene details, motion patterns, and plot elements. All questions are formulated so their answers can be directly inferred from the video.
\textbf{3) QA-based Scoring.}
Given the generated caption $\hat{T}$ and question set $\{Q_i\}_{i=1}^N$, an LLM produces predicted answers $\{\hat{A}_i\}$. These are compared with ground-truth answers $\{A_i^\ast\}$ derived from the condition. Each question receives a binary correctness score
$s_i = \mathbb{1}(\hat{A}_i = A_i^\ast)$, and the final alignment score is $S_{\text{align}} = \frac{1}{N}\sum_{i=1}^N s_i \in [0,1]$.
A higher score indicates stronger consistency between the generated video and the input condition. More details about prompt and questions can be seen in \textbf{supplementary materials}.

\subsection{4D Consistency}

To comprehensively evaluate the spatial-temporal stability of generated videos, we assess the \textbf{4D Consistency} from three complementary perspectives: \textbf{3D Consistency}, \textbf{Motion Consistency}, and \textbf{Style Consistency}. These three metrics jointly quantify the geometric, dynamic, and perceptual coherence of the generated scene sequences.

\noindent \textbf{3D Consistency.}
We measure geometric consistency as the reprojection error of 3D points reconstructed by a dense, differentiable SLAM method~\cite{teed2021droid,schonberger2016structure,duan2025worldscore}. To reduce sensitivity to FPS and camera motion, we compute this error on short temporal clips and average over all clips to obtain the final consistency score for both 3D and 4D videos.

\noindent \textbf{Motion Consistency.}
We evaluate motion consistency with two complementary components: a feature-based optical-flow similarity metric and an MLLM-based judge. For each consecutive frame pair, we estimate optical flow and compare it with the flow  induced by the generated motion field.
 In parallel, to prevent certain videos from exhibiting minor unintended motions, we apply MLLM-as-Judge as described in Sec. 4.2 to perform yes/no question–answering over each video. It evaluates object-level motion rationality (e.g., continuity, interactions, speed and trajectory plausibility), and we use the mean correctness as the motion rationality score. Together, this 
\textbf{hybrid evaluation method} jointly assesses low-level temporal coherence and high-level semantic consistency of motion.

\noindent \textbf{Style Consistency.}
We assess style consistency ~\cite{duan2025worldscore} by comparing the Gram matrices of deep features between frames. To reduce sensitivity to video length and FPS, we further compute this metric over short temporal clips and average the results as the final style consistency score.
More details about prompt and questions can be seen in \textbf{supplementary materials}.

\subsection{Perceptual Quality}

To further evaluate the perceptual realism of generated videos, we measure four complementary aspects of perceptual quality.
\noindent\textbf{1) Spatial Quality.} 
We assess the spatial fidelity and technical quality of individual frames using CLIPIQA+~\cite{wang2023exploring}. 
The overall aesthetic appeal of generated frames is evaluated using CLIP-Aesthetic~\cite{schuhmann2022laion}, which estimates human aesthetic preferences based on CLIP embeddings.
We combine the outputs of these two models to obtain the overall spatial quality score.
\noindent\textbf{2) Temporal Quality.}
To evaluate temporal coherence and visual stability across frames, we employ FastVQA~\cite{wu2022fast}, a video quality assessment model designed for efficient temporal perceptual evaluation.
\noindent\textbf{3) 3D Texture Quality.}
We further measure 3D texture realism by prompting mPLUG-Owl3~\cite{ye2024mplug} to provide ratings on 3D texture fidelity. 
More details about prompt and questions can be seen in \textbf{supplementary materials}.

\begin{table*}[htbp]
\centering
\scriptsize
\caption{Leaderboard for different categories of 3D generation models.}
\vspace{-3mm}
\resizebox{0.8\linewidth}{!}{\begin{tabular}{l|ccc|ccc|cc|c}
\toprule
\multirow{2}{*}{\textbf{Model}} 
& \multicolumn{3}{c|}{\textbf{Perceptual Quality}} 
& \multicolumn{3}{c|}{\textbf{Condition-3D Alignment}} 
& \multicolumn{2}{c|}{\textbf{3D Consistency}} 
& \multirow{2}{*}{\textbf{Overall}} \\ 
\cmidrule(lr){2-4} \cmidrule(lr){5-7} \cmidrule(lr){8-9} 
& Spatial & Temporal & 3D Texture 
 & Scene & Attribute & Relationship 

& Viewpoint & Style 
& \\ 
\midrule

\multicolumn{10}{l}{\textbf{Image-to-3D}} \\ 
SyncDreamer~\cite{liu2023syncdreamer} (Obj) & 0.498+0.480 &0.460 &0.622  &0.772 &0.864&0.895 & 0.117&0.943 &0.628  \\ 
V3D~\cite{chen2024v3d} (Obj)          & 0.496+0.439& 0.669 &0.670 & 0.788& 0.920&0.872  & 0.470  &0.904 &0.692\\ 
MotionCtrl~\cite{wang2024motionctrl} (Sce)   &0.424+0.508 &0.738 &0.710& 0.927 &0.926&0.837  &0.970  &0.888 &0.770  \\ 
Viewcrafter~\cite{yu2024viewcrafter} (Sce)  &0.497+0.552 &0.797 & 0.715  & 0.972 &0.954&0.863  & 0.801 &0.698 &0.761  \\ 
FlexWorld~\cite{chen2025flexworld} (Sce)    &0.437+0.508 &0.745 & 0.701&  0.944& 0.953&0.813& 0.931& 0.780&0.757  \\ 

\midrule
\multicolumn{10}{l}{\textbf{Text-to-3D}} \\ 
Director3D~\cite{li2024director3d}    & 0.180+0.511 &0.601 &0.677 &0.737 &0.813&0.541  &0.991 &0.992 &0.671  \\ 
Text2NeRF~\cite{zhang2024text2nerf}      &0.271+0.514 &0.754 &0.677 & 0.798& 0.805&0.605  &0.988 &0.878 &0.699 \\ 
Step1x-to-3D~\cite{li2025step1x}   &0.455+0.387& 0.828 & 0.596 &0.423 & 0.605&0.490 & 0.806 &0.971 &0.618 \\ 
WonderJourney~\cite{yu2024wonderjourney}  & 0.284+0.449 &0.657 &0.703 &0.821& 0.855&0.683& 0.774 & 0.602 &0.619 \\ 


\bottomrule
\end{tabular}}
\vspace{-3mm}
\end{table*}

\begin{table}[t]
\centering
\small
\setlength{\tabcolsep}{6pt}
\caption{Performance on Videophy2-test. PC means physical commonsens, SA means semantic adherence, ranging from 0 to 5. The Min(PC,SA) returns the smaller of PC and SA, the Joint is binary (0 or 1), assigned 1 only when both PC and SA larger than 4. }
\resizebox{0.98\linewidth}{!}{\begin{tabular}{lcccc}
\toprule
Method & Min(PC,SA) ACC/F1  & Joint ACC/F1 & Min(PC,SA) PLCC/SRCC  & Joint PLCC/SRCC \\
\midrule
Ours & \textbf{0.469/0.439} & 0.677/0.631  & \textbf{0.408/0.409} & \textbf{0.369/0.376}\\
Videophy2        & 0.450/0.452 & \textbf{0.770/0.656}  & 0.394/0.385 & 0.308/0.308\\
PhygenBench       & 0.220/0.181 & 0.360/0.359  & 0.050/0.066  & 0.009/0.029 \\
Vbench2          & 0.220/0.123 & 0.223/0.187 & 0.240/0.243 & 0.225/0.235 \\
\bottomrule
\end{tabular}}
\vspace{-5mm}
\label{tab:acc_f1_vbench2}
\end{table}

\begin{figure*}[tbp]
  \centering
   \includegraphics[width=0.8\linewidth]{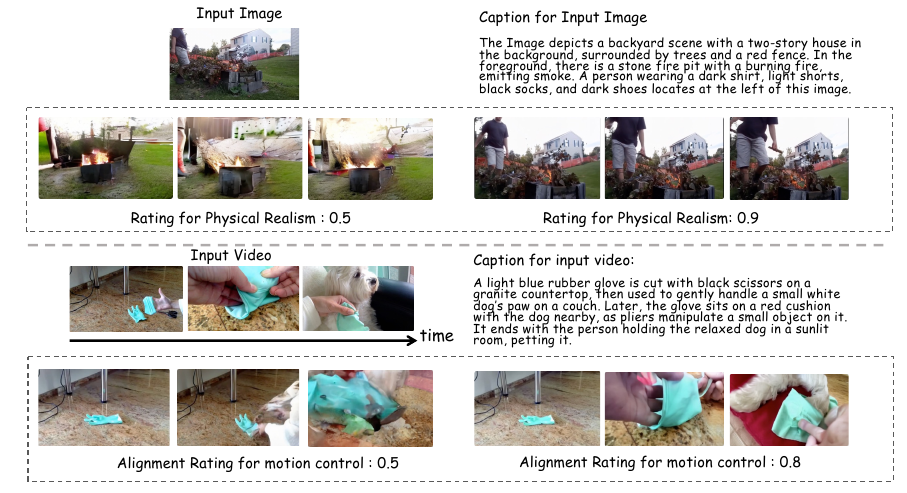}
   \vspace{-3mm} 
   \caption{Visualization of some rating examples for physical realism and motion control (4D-Condition Alignment).}
   \vspace{-3mm} 
   \label{fig:demo_visual}
\end{figure*}
\section{Experiment}

\subsection{Experiment Details}
\noindent \textbf{Evaluated Metrics:} Most of the metrics correspond to standalone methodologies. For perceptual quality, our evaluation integrates technical quality and aesthetic quality (as reflected by the a + b components in the table). For motion consistency, we employ a combination of feature-based similarity and MLLM-based QA (corresponding to the a + b components in the table). 
\noindent \textbf{Evaluated Models:}
For \textbf{Image-to-3D}, we select SyncDreamer~\cite{liu2023syncdreamer}, V3D~\cite{chen2024v3d}, MotionCtrl~\cite{wang2024motionctrl}, ViewCrafter~\cite{yu2024viewcrafter}, FlexWorld~\cite{chen2025flexworld}.
For \textbf{Image-to-4D}, we select CamI2V~\cite{zheng2024cami2v} and Diffusion-as-Shader~\cite{gu2025diffusion}.
For \textbf{Text-to-3D}, we select Director3D~\cite{li2024director3d}, Text2NeRF~\cite{zhang2024text2nerf}, Step1X-3D~\cite{li2025step1x}, WonderJourney~\cite{yu2024wonderjourney}.
For \textbf{Text-to-4D}, we select 4Dfy~\cite{4dfy} and dreamin4D~\cite{dreamin4D}
For \textbf{Video-to-4D}, we select ReCamMaster~\cite{bai2025recammaster}, TrajectorCrafter~\cite{yu2025trajectorycrafter}, EX-4D~\cite{ex4d}, and Vista~\cite{gao2024vista}. Because the 3D model contains no object motion, we omit evaluations of event control, motion control, physical realism, and motion consistency, but since object rotation is presented as a video, we still assess perceptual temporal quality.

\subsection{Results and Discussion on Physical Realism} 

\noindent \textbf{Video-conditioned models.} 
Video-to-4D models achieve the excellent performance in physical realism, particularly in modeling dynamic behavior. ReCamMaster~\cite{bai2025recammaster} and TrajectoryCrafter~\cite{yu2025trajectorycrafter} lead the dynamics sub-metric with scores of (0.680, 0.714, 0.773) and (0.671, 0.667, 0.636), respectively. They also show high optical realism (0.714 and 0.667), benefiting from the spatiotemporal continuity of video inputs. Despite EX-4D~\cite{ex4d} having lower dynamics (0.421), it still outperforms text-based models, indicating that even weaker video baselines preserve more physical plausibility than text-conditioned models.
\noindent \textbf{Image-conditioned models.} 
In Image-to-4D models, DiffusionAsShader~\cite{gu2025diffusion} demonstrate superior physical realism on Optics, Thermal and a strong Dynamics score. \textbf{It also outperforms other modality-conditioned models}.
\noindent \textbf{Text-conditioned models.} 
Text-to-4D models exhibit the weakest performance across all physical realism metrics. Even the stronger model, 4Dfy~\cite{4dfy}, achieves only 0.265 in dynamics and 0.417 in optics, with a low thermal score of 0.200. dreamin4D~\cite{dreamin4D} performs even lower, particularly in optics (0.238), revealing the inherent difficulty of generating physically plausible scenes solely from text. These results emphasize the modality gap in physical realism, with language-conditioned generation still lacking the inductive bias necessary for faithful physical simulation.
The rating visualization for physical realism can be seen in Fig.~\ref{fig:demo_visual}.


\subsection{Results and Discussion on 4D Consistency}

\noindent \textbf{Image-to-3D models}: MotionCtrl~\cite{wang2024motionctrl} and FlexWorld~\cite{chen2025flexworld} achieve high viewpoint consistency (0.970 and 0.931), with strong style retention as well (0.888 and 0.780, respectively). V3D~\cite{chen2024v3d} scores moderately (viewpoint: 0.470, style: 0.904), while SyncDreamer~\cite{liu2023syncdreamer} performs relatively poorly in viewpoint (0.117) but retains high style (0.943), suggesting potential instability in geometric alignment but success in style consistency.
\noindent \textbf{Text-to-3D models}: Director3D~\cite{li2024director3d} and Text2NeRF~\cite{zhang2024text2nerf} both reach high viewpoint consistency (0.991 and 0.988), indicating strong camera-invariant reconstruction from language. Style-wise, Director3D (0.992) again leads. Step1x-to-3D balances moderate viewpoint (0.806) with good style (0.971), indicating varying trade-offs across architecture.
\noindent \textbf{Text-to-4D models}: 4Dfy~\cite{4dfy} consistently outperforms dreamin4D~\cite{dreamin4D} in all three sub-metrics. It achieves especially high performance in motion (0.934+0.705), viewpoint consistency (0.741) and style consistency (0.993), suggesting better viewpoint change and temporal change consistency despite the input modality limitations. 
\noindent \textbf{Image-to-4D models}: DiffusionAsShader~\cite{gu2025diffusion} surpasses CamI2V~\cite{zheng2024cami2v} across all metrics, particularly in motion (0.874+0.750 vs. 0.553+0.816) and viewpoint (0.908 vs. 0.701). Style consistency also favors DiffusionAsShader (0.918 vs. 0.887), indicating better 4D consistency. This suggests that 3D-aware diffusion frameworks may maintain 4D consistency more effectively than 2D video-diffusion methods with geometric priors.
\noindent \textbf{Video-to-4D models}: ReCamMaster~\cite{bai2025recammaster} achieves the highest overall consistency, with excellent performance in all three areas (viewpoint: 0.862, motion: 0.859+0.834, style: 0.985). In contrast, EX-4D~\cite{ex4d} and TrajectoryCrafter~\cite{yu2025trajectorycrafter} show lower viewpoint and motion coherence, despite moderate performance in style. Notably, Vista~\cite{gao2024vista} is for autonomous driving, and while it achieves strong consistency across feature-based metrics (0.942) due to minimal object movement, its performance degrades (0.622) on semantic QA where dynamic object movement is required.

\vspace{-1mm}
\subsection{Results on Condition-3D/4D Alignment}
\vspace{-1mm}
\noindent \textbf{Image-3D models}: 
Viewcrafter~\cite{yu2024viewcrafter} achieves the best overall condition alignment, especially in scene (0.972) and attribute alignment (0.954). 
MotionCtrl~\cite{wang2024motionctrl} and FlexWorld~\cite{chen2025flexworld} also perform well in attribue and relationship-aligned generation (both large than 0.92). SyncDreamer~\cite{liu2023syncdreamer} and V3D~\cite{chen2024v3d} show weaker spatial scene understanding and SyncDreamer has lower attribute alignment.
\noindent \textbf{Image-4D models}:
DiffusionAsShader~\cite{gu2025diffusion} outperforms CamI2V~\cite{zheng2024cami2v}, achieving higher motion control (0.781), scene control (0.965) and event control (0.4), as well as top attribute alignment (0.963), demonstrating strong fidelity to both static and relational visual cues from input images.
\noindent \textbf{Video-4D models}: 
TrajectoryCrafter~\cite{yu2025trajectorycrafter} leads in overall condition alignment within video-to-4D models, especially excelling in motion (0.856) and scene (0.757) alignment. Vista has the lowest attribute alignment (0.516), likely due to minimal object movement. EX-4D~\cite{ex4d} demonstrates a good balance across different 4D consistency.
\noindent \textbf{Text-3D models}: 
WonderJourney~\cite{yu2024wonderjourney} outperforms other models in scene (0.821) and attribute alignment (0.855), suggesting better adherence to textual semantics. 
\noindent \textbf{Text-to-4D models}: 4Dfy~\cite{4dfy} and dreamin4D~\cite{dreamin4D} show modest attribute alignment (0.326 and 0.390), consistent with their generation style being limited to object-centric turntable rotations. Text-conditioned models exhibit larger gaps in motion control due to weak temporal grounding.
The rating visualization for motion control can be seen in Fig.~\ref{fig:demo_visual}. More visualization examples can be seen in \textbf{supplementary materials}.
\vspace{-1mm}
\subsection{Perceptual Quality Analysis}
\vspace{-1mm}
\textbf{To-3D models:}
Among 3D generation models, image-conditioned approaches such as Viewcrafter~\cite{yu2024viewcrafter} and MotionCtrl~\cite{wang2024motionctrl} achieve the best overall perceptual quality,  while text-conditioned ones like Step1x-to-3D~\cite{li2025step1x} exhibit better temporal coherence but weaker spatial fidelity. Overall, image-to-3D models deliver higher spatial and texture quality.
\textbf{To-4D models:}
For 4D generation, DiffusionAsShader (image-conditioned)~\cite{gu2025diffusion} attains the highest perceptual realism and temporal smoothness, followed by TrajectoryCrafter~\cite{yu2025trajectorycrafter} (video-conditioned) with superior motion coherence, and 4Dfy~\cite{4dfy} (text-conditioned) showing modest but improving spatio-temporal stability. Image and video inputs thus favor structure and dynamics.

\subsection{Ablation for Metric Design}
\vspace{-1mm}
\textbf{Key-points on Physical Consistency Evaluation.}
We conducted a subjective evaluation involving around 15 participants on 100 videos to construct a physical-assessment test set, on which we further performed ablation experiments to analyze the metric improvements. The performance comparison of different methods on this subjective test set is presented in Table~\ref{tab:physics_eval_abla}.
The results in Table~\ref{tab:physics_eval_abla} highlight three key findings.
First, our LLM-as-judge design consistently outperforms MLLM-based judging, demonstrating the advantage of text-driven reasoning for physical assessment.
Second, using more diagnostic questions leads to stronger alignment with human judgment, indicating that richer questioning improves evaluation reliability.
Third, adaptive dimension selection (AdaDimen) yields clear gains over fixed dimension settings, showing that LLMs are more effective when allowed to identify semantically relevant physical dimensions rather than relying on predefined ones.

\noindent\textbf{Comparison on Physical Consistency Evaluation.}
We evaluate our method on the physical-consistency benchmarks covering Physical Commonsense  (PC), Semantic Adherence (SA), their conservative minimum of PC \& SA, and a binary joint score. Across all evaluation settings, our training-free method achieves performance comparable to—or in some cases exceeding—the trained VideoPhy2~\cite{bansal2025videophy} model, while substantially outperforming other training-free baselines. This demonstrates that our approach provides robust and reliable physical reasoning without requiring task-specific training.
For more metric ablation studies, please see the \textbf{supplementary materials}.

\begin{table}[t]
\centering
\caption{Comparison of judge types, dimension settings, and question counts on Physical Realism Evaluation. ``FixDimen'' uses predefined dimensions; ``AdaDimen'' applies adaptive dimension.}
\resizebox{\linewidth}{!}{
\begin{tabular}{lccccl}
\toprule
\textbf{Setting} & \textbf{Judge (Model)} & \textbf{Dim Type} & \textbf{\#Ques} & \textbf{PLCC / SRCC} \\
\midrule
\textbf{VBench2} & MLLM (Llava-Video) & FixDimen & 2 & 0.341 / 0.379 \\
\textbf{Ours-v1} & LLM(GPT-5$\rightarrow$GPT-5) & FixDimen & 10 & 0.351 / 0.402 \\
\textbf{Ours-v2} & LLM (GPT-5$\rightarrow$GPT-5) & AdaDimen & 10 & \textbf{0.452 / 0.461} \\
\textbf{Ours} & Hybrid(GPT-5$\rightarrow$Qwen2.5-VL) & AdaDimen & 10 & 0.247 / 0.237\\

\bottomrule
\end{tabular}}
\vspace{-3mm}
\label{tab:physics_eval_abla}
\end{table}

\section{Conclusion}

In this work, we introduced \textbf{4DWorldBench}, a unified, multimodal, and physics-aware benchmark for evaluating next-generation world generation models. Unlike prior video-centric or physics-specific evaluations, 4DWorldBench jointly measures \emph{perceptual quality}, \emph{condition--4D alignment}, \emph{physical realism}, and \emph{4D consistency} within a single, extensible framework. By integrating both MLLM- and LLM-based Question Answering (QA) evaluators, our adaptive methodology enables fine-grained, semantically grounded assessment across diverse text, image, and video conditions. And we also apply hybrid evaluation methods for some dimensions. We hope 4DWorldBench can serve as a standardized platform for benchmarking world generation, promoting fair comparison, deeper understanding, and future innovation toward more coherent, controllable, and physically consistent virtual worlds. In future work, we plan to extend the benchmark to broader real-world scenarios and investigate lightweight evaluation paradigms to further improve scalability and accessibility.

{
    \small
    \bibliographystyle{ieeenat_fullname}
    \bibliography{main}
}


\maketitlesupplementary

\section{More Details for Benchmark Metrics}

\subsection{Physical Realism}
The prompts for evaluating physical realism in both question–answer generation and caption-based reasoning are presented as follows.

\begin{promptbox}{Physics Reasoning QA Template}

\textbf{Video Caption:} \textless{}Video Caption\textgreater{}

\vspace{1em}

Above is the caption of a video. Please generate yes/no questions for evaluating whether the described scenario follows real-world physics.

\textbf{Instruction:} You are a scientist who designs diagnostic yes/no questions about short real-world scenarios for physics evaluation.

Given: (1) a natural-language description of a real-world scene, and (2) an ordered list of dimensions (e.g., Fundamental Physics, Optics, Material Interaction \& Transformation, Force \& Motion, Thermal \& Phase Transition, etc.).

Returns ten (10) questions in the order provided, with one to four questions per dimension. If there are too few dimensions to ask, more questions per dimension will be required.

Reasoning about reasonable phenomena that should occur in the real world based on the description and posing them as questions, such as a balloon should explode when a sharp object presses into it, water should boil at above 100 degrees, cheese should melt at high temperatures, etc.

You should reason about what should happen in the real world, each question should be crafted to be answerable solely by inspecting the description and focus on visible phenomenon in the description without requiring external knowledge.

If the input description does not contain some problem dimensions or cannot design “yes” answer questions, skip generating questions for that part and move on to the next dimension and do not invent imaginary properties.

Each question must stay strictly within its assigned dimension's scope. Avoid cross-dimension leakage.

Use present tense, neutral tone, and end each question with “(yes or no)”.

Return a JSON array of objects, each with:

\begin{quote}
\{ 'dimension': '\textless dimension name\textgreater',  
'auxiliary\_info': ['\textless one to four yes/no questions\textgreater'] \}
\end{quote}

Preserve the dimension order. Validate: at most 8 objects, each auxiliary\_info has one to four questions, all questions are dimension-appropriate, observable, and answer 'yes'.

Here are some in-context examples:

\begin{quote}
'questions': [ \\
\{ 'dimension': 'Material Interaction \& Transformation', \\
\quad 'auxiliary\_info': [ \\
\quad 'Does the blue and yellow paints visibly exist? (yes or no)', \\
\quad 'Does the blue and yellow paints mix visibly during the stirring process? (yes or no)', \\
\quad 'Does the blue and yellow disappear from the mixed paint? (yes or no)', \\
\quad 'Finally, does the paint become green? (yes or no)' ] \}, \\

\{ 'dimension': 'Force \& Motion', \\
\quad 'auxiliary\_info': [ \\
\quad 'Does the toothpaste tube contact with hands? (yes or no)', \\
\quad 'Does the toothpaste tube deform under stress? (yes or no)', \\
\quad 'Does the toothpaste tube be compressed and the toothpaste be expelled out of the toothpaste tube? (yes or no)' ] \}, \\

\{ 'dimension': 'Thermal \& Phase Transition', \\
\quad 'auxiliary\_info': [ \\
\quad 'Initially, is the river in a liquid state? (yes or no)', \\
\quad 'Finally, does the river freeze? (yes or no)' ] \} ]
\end{quote}
\end{promptbox}

\begin{promptbox}{System Prompt for Physics Question Generation}
\textbf{Role:} JSON-only assistant for physics question generation.

\medskip
\textbf{System Prompt:}
\begin{itemize}[leftmargin=*,nosep]
  \item ``You are an useful assistant that only outputs valid JSON format. Always use double quotes for keys and values, and never use single quotes or any extra text.The format should be: questions:[q1,q2,...].''
\end{itemize}
\end{promptbox}

\begin{promptbox}{Physics Question Design Template}
\textbf{Scene Description:} \textit{\textless{}Description\textgreater{}}

\medskip
Above is a natural-language description of a real-world scene and an
ordered list of physics-related dimensions. Please design diagnostic
yes/no questions for physics evaluation.

\medskip
\textbf{Instruction (Physics Reasoning QA Template):}
\begin{enumerate}[leftmargin=*,nosep]
  \item You are a scientist who designs diagnostic yes/no questions about
        short real-world scenarios.
  \item Given: (1) a natural-language description of a real-world scene,
        and (2) an ordered list of dimensions (e.g., Fundamental Physics,
        Optics, Material Interaction \& Transformation, Force \& Motion,
        Thermal \& Phase Transition, etc.).
  \item Return ten (10) questions in the order provided, with one to four
        questions per dimension. If there are too few dimensions to ask,
        more questions per dimension will be required.
  \item Reason about reasonable phenomena that should occur in the real
        world based on the description and pose them as questions (e.g., a
        balloon should explode when a sharp object presses into it, water
        should boil above 100 degrees, cheese should melt at high
        temperatures, etc.).
  \item Each question should be crafted to be answerable solely by
        inspecting the description and focus on visible phenomena in the
        description, without requiring external knowledge.
  \item If the input description does not contain some dimensions or you
        cannot design a question whose answer is ``yes'', skip that
        dimension and do not invent imaginary properties.
  \item Each question must stay strictly within its assigned dimension's
        scope and avoid cross-dimension leakage.
  \item Use present tense and neutral tone, and end each question with
        ``(yes or no)''.
  \item \textbf{Output format:} Return a JSON array of objects, each with:
        \texttt{\{ "dimension": "<dimension name>", "auxiliary\_info":
        ["<one to four yes/no questions>"] \}}.
  \item Preserve the dimension order. Validate: at most 8 objects, each
        \texttt{auxiliary\_info} has one to four questions, all questions
        are dimension-appropriate, observable, and answer ``yes''.
\end{enumerate}

\medskip
\textbf{In-context examples:} The prompt also includes example objects for
\emph{Material Interaction \& Transformation}, \emph{Force \& Motion}, and
\emph{Thermal \& Phase Transition}, each with several yes/no questions
ending with ``(yes or no)''.
\end{promptbox}

\begin{promptbox}{System Prompt for Physics Answer Evaluation}
\textbf{Role:} JSON-only assistant for answering physics questions.

\medskip
\textbf{System Prompt:}
\begin{itemize}[leftmargin=*,nosep]
  \item ``You are an assistant that only outputs valid JSON format. Always use double quotes for keys and values, and never use single quotes or any extra text. Example: {"answer":"yes"} or {"answer":"no"}''
\end{itemize}
\end{promptbox}

\begin{promptbox}{Caption-based Physics Answer Template}
\textbf{Video Caption:} \textit{\textless{}Caption\textgreater{}}

\medskip
\textbf{Question:} \textit{\textless{}Physics Question\textgreater{}}

\medskip
\textbf{Instruction:}  
You are an expert at answering questions based on descriptions of generated
videos, which may contain various physically unreasonable. Please answer
\textbf{yes or no only} for the following question according to the
caption. When answering, you should carefully check whether the main
objects and behaviors of the question and caption are consistent.

\medskip
The model must output JSON in the form
\texttt{\{"answer":"yes"\}} or \texttt{\{"answer":"no"\}}.
\end{promptbox}

\subsection{Condition-4D Alignment}

\begin{figure*}[tbp]
  \centering
   \includegraphics[width=0.9\linewidth]{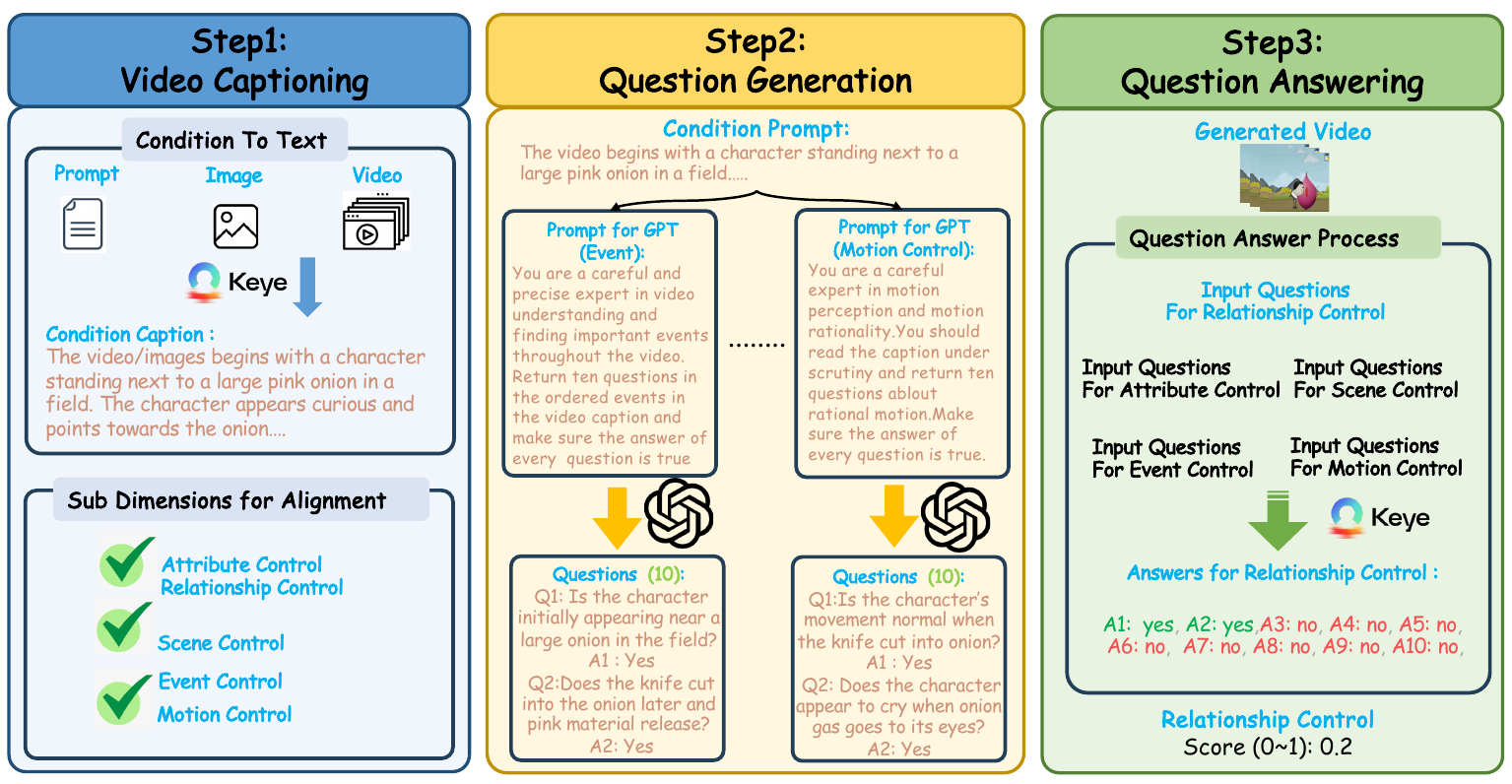}
   \vspace{-3mm} 
   \caption{Overall pipeline of 4D-Condition Alignment Evaluation. The framework converts multimodal conditions into text, generates fine-grained and dimension-specific questions for sub-aspects, and uses an MLLM-based QA process to assess the alignment score}
   \vspace{-3mm} 
   \label{fig:align}
\end{figure*}

\paragraph{Main Framework}
For Condition-4D Alignment, it follows a systematic three-step process—condition input captioning, question generation, and question answering—to quantitatively evaluate alignment quality, as illustrated in Fig.~\ref{fig:align}. Unlike the physical-realism track, which focuses on whether the model can faithfully simulate or reason about complex physical behaviors in 4D space, Condition-4D Alignment targets a different dimension: whether the model can maintain coherent, semantically accurate alignment between user-specified conditions and the generated video content. This dual-track design is motivated by the observation that current Multimodal Large Language Models (MLLM), while proficient at surface-level video question answering, often struggle with complex physical reasoning (e.g., fluid dynamics, force interactions), whereas LLMs excel at abstract reasoning and compositional understanding when provided with high-quality textual descriptions. By decoupling physical realism from condition-driven semantic alignment, our evaluation isolates complementary capabilities and enables a more complete diagnosis of model performance.
\paragraph{Camera Control.} 
Following WorldScore~\cite{duan2025worldscore}, we evaluate camera controllability by comparing the reconstructed camera trajectory with the ground-truth control trajectory. For each generated video, we first estimate frame-wise camera poses using DROID-SLAM~\cite{teed2021droid}. We then measure the angular deviation between the ground-truth and estimated rotations (in degrees) as
\begin{equation}
e_{\theta} = \arccos \left( \frac{\mathrm{tr}(\mathbf{R}_{\text{gt}}\mathbf{R}^\top) - 1}{2} \right) \cdot \frac{180}{\pi},
\end{equation}
and the scale-invariant Euclidean distance between the ground-truth and estimated camera centers as
\begin{equation}
e_{t} = \big\|\mathbf{t}_{\text{gt}} - s \mathbf{t}\big\|_2,
\end{equation}
where $\mathbf{R}_{\text{gt}}, \mathbf{R} \in \mathrm{SO}(3)$ are the ground-truth and estimated rotation matrices, $\mathbf{t}_{\text{gt}}, \mathbf{t} \in \mathbb{R}^3$ are the corresponding camera positions, and $s$ is the least-squares scale factor. We combine the rotational and translational errors using the geometric mean to obtain a per-frame camera error
\begin{equation}
e_{\text{camera}} = \sqrt{e_{\theta} \cdot e_{t}}.
\end{equation}
The final camera controllability error for a model is obtained by averaging $e_{\text{camera}}$ over all frames of all generated videos, where lower values indicate better adherence to the desired camera trajectory.
Note that, unlike other parts of our benchmark, we do not rely on MLLM-based question–answering here, since current multimodal language models exhibit limited ability to accurately reason about fine-grained camera motions.

\paragraph{Prompts for MLLM QA} The prompts for evaluating 4D-Condition Alignment as follows.

\begin{promptbox}{System Prompt for QA JSON Output}
\textbf{Role:} JSON-only assistant for QA generation.

\medskip
\textbf{System Prompt:}
\begin{itemize}[leftmargin=*,nosep]
  \item ``You are an assistant that only outputs valid JSON format. Return the questions as a JSON array of strings.''
\end{itemize}
\end{promptbox}

\begin{promptbox}{QA Template for Spatial Relationship Control Evaluation}

Please generate detailed yes/no questions
about spatial relationships and relative position changes of objects over
time based on this caption.

\medskip
\textbf{Instruction:} You are an expert in caption analysis focusing on object
spatial relationship and relative position changes in the whole caption.

\medskip
\textbf{Note:}
\begin{enumerate}[leftmargin=*,nosep]
  \item Analyze the following video content and generate 5 specific yes/no
        questions that evaluate the spatial relationship between objects and
        their relative position changes over time.
  \item The description of the video is:

        \textit{Video content: \{content\}}
  \item Requirements:
  \begin{enumerate}[leftmargin=*,nosep,label*=\arabic*.]
    \item Generate exactly 5 questions.
    \item Each question should be answerable with yes/no and the answer of
          every question should be \textbf{yes}.
    \item Focus on spatial relationships and relative position changes.
    \item Questions should be specific to the video content.
    \item The output should be a JSON list of strings.
  \end{enumerate}
\end{enumerate}
\end{promptbox}

\begin{promptbox}{ QA Template for Attribute Control Evaluation}

\medskip
Please generate detailed yes/no questions
about dynamic attributes and object transformations.

\medskip
\textbf{Instruction:} You are an expert in video analysis focusing on dynamic
attributes and object transformations.

\medskip
\textbf{Note:}
\begin{enumerate}[leftmargin=*,nosep]
  \item Analyze the following video content and generate 5 specific yes/no
        questions that evaluate whether objects in the video show dynamic
        changes in their attributes (color, size, shape, texture, state, etc.).
  \item The description of the video is:

        \textit{Video content: \{content\}}
  \item Requirements:
  \begin{enumerate}[leftmargin=*,nosep,label*=\arabic*.]
    \item Generate exactly 5 questions.
    \item Each question should be answerable with yes/no.
    \item Focus on temporal changes and object transformations.
    \item Questions should be specific to the video content.
    \item The output should be a JSON list of strings.
  \end{enumerate}
\end{enumerate}
\end{promptbox}

\begin{promptbox}{QA Template for Event Control}
\medskip
 Please generate
yes/no questions that evaluate the story plot, character actions, and
narrative progression in chronological order.

\medskip
\textbf{Instruction:} You are an expert in story analysis and event
understanding.

\medskip
\textbf{Note:}
\begin{enumerate}[leftmargin=*,nosep]
  \item Analyze the following video content and generate 10 specific yes/no
        questions that evaluate the event, story plot, character actions, and
        narrative progression in chronological order.
  \item The description of the video is:

        \textit{Video content: \{content\}}
  \item Requirements:
  \begin{enumerate}[leftmargin=*,nosep,label*=\arabic*.]
    \item Generate exactly 10 questions.
    \item Each question should be answerable with yes/no.
    \item Questions should follow the chronological order of events.
    \item Focus on story elements, character actions, and plot development.
    \item Questions should be specific to the video content.
    \item The output should be a JSON list of strings.
  \end{enumerate}
\end{enumerate}
\end{promptbox}

\begin{promptbox}{QA Template for Complex Scene Control}

Please generate yes/no questions about detailed scene content in temporal
order.

\medskip
\textbf{Instruction:} You are an expert in scene analysis focusing on
complex scene and environments.

\medskip
\textbf{Note:}
\begin{enumerate}[leftmargin=*,nosep]
  \item Analyze the following video caption and generate 10 specific yes/no
        questions that evaluate the detailed content of the landscape and
        environment of the video caption in time order.
  \item The description of the video is:

        \textit{Video content: \{content\}}
  \item Requirements:
  \begin{enumerate}[leftmargin=*,nosep,label*=\arabic*.]
    \item Generate exactly 10 questions.
    \item Each question should be answerable with yes/no and the answer of
          every question should be \textbf{yes}.
    \item Focus on detailed landscape and environment elements.
    \item Raise questions about the landscape and scene content of the video
          caption in time order.
    \item The output should be a JSON list of strings.
  \end{enumerate}
\end{enumerate}
\end{promptbox}

\begin{promptbox}{QA Template for Motion Control}
 
Please generate
yes/no questions about the temporal order and sequence of motions.

\medskip
\textbf{Instruction:} You are an expert in motion analysis and temporal
motion sequence understanding.

\medskip
\textbf{Note:}
\begin{enumerate}[leftmargin=*,nosep]
  \item Analyze the following video caption and generate 10 specific yes/no
        questions that evaluate the temporal order and sequence of motions
        described in the video caption.
  \item The description of the video is:

        \textit{Video content: \{content\}}
  \item Requirements:
  \begin{enumerate}[leftmargin=*,nosep,label*=\arabic*.]
    \item Generate exactly 10 questions.
    \item Each question should be answerable with yes/no and the answer of
          every question should be \textbf{yes}.
    \item Focus on temporal order and sequence of motions.
    \item Raise specific questions about the existing motions in the video to
          validate whether the motions in the video are consistent with those
          described in the caption in time order.
    \item The output should be a JSON list of strings.
  \end{enumerate}
\end{enumerate}
\end{promptbox}

\subsection{4D Consistency}

To comprehensively evaluate the spatial-temporal stability of generated videos, we assess the \textbf{4D Consistency} from three complementary perspectives: \textbf{3D Consistency}, \textbf{Motion Consistency}, and \textbf{Style Consistency}. These three metrics jointly quantify the geometric, dynamic, and perceptual coherence of the generated scene sequences.

\noindent \textbf{3D Consistency.}
We measure geometric consistency using the reprojection error of 3D points reconstructed by a dense, differentiable SLAM pipeline~\cite{teed2021droid,schonberger2016structure,duan2025worldscore}. For each temporal clip $c \in \mathcal{C}$, the clip-level reprojection error is
\begin{equation}
e_{\text{reproj}}^{(c)} =
\frac{1}{|\mathcal{V}c|}
\sum{(i,j)\in\mathcal{V}c}
\left|
\mathbf{p}^{*}{ij} - \Pi(\mathbf{P}_{ij})
\right|_2,
\end{equation}
where $\mathcal{V}c$ is the set of co-visible pixels inside clip $c$, $\mathbf{p}^{*}{ij}$ is the observed pixel, $\mathbf{P}_{ij}$ is the reconstructed 3D point, and $\Pi(\cdot)$ is the projection operator.
The final 3D consistency score is obtained by averaging over all clips:
\begin{equation}
e_{\text{3D}} = 1-nomarlize(\frac{1}{|\mathcal{C}|} \sum_{c \in \mathcal{C}} e_{\text{reproj}}^{(c)}).
\end{equation}
\paragraph{Motion Consistency.}
We evaluate motion consistency using a flow-based temporal smoothness metric~\cite{duan2025worldscore} and an MLLM-based motion rationality score.
For each clip $c$ with $T_c$ frames, the flow error is
\begin{equation}
e_{\text{flow}}^{(c)} =
\frac{1}{T_c - 1}
\sum_{t=1}^{T_c - 1}
\left|
\mathbf{F}{t\rightarrow t+1} -
\mathbf{F}'{t\rightarrow t+1}
\right|_2,
\end{equation}
where $\mathbf{F}{t\rightarrow t+1}$ is the estimated optical flow and $\mathbf{F}'{t\rightarrow t+1}$ is the flow induced by the predicted motion field.
In parallel, an MLLM performs video-level yes/no question–answering to assess object-level motion rationality (continuity, interactions, and trajectory plausibility). Let $s_{\text{QA}}^{(c)} \in [0,1]$ denote the mean correctness of its answers for clip $c$.
The final motion consistency score averages the two components:
\begin{equation}
e_{\text{motion}}=\left(1- normalize(
\frac{1}{|\mathcal{C}|}
\sum_{c \in \mathcal{C}}
e_{\text{flow}}^{(c)} )\right) + s_{\text{QA}}
,
\end{equation}
where a higher score indicates better temporal coherence and semantic rationality.

\paragraph{Prompt} The prompts for evaluating Motion Consistency as follows.
\begin{promptbox}{ QA Template for Motion Rationality Evaluation in Video Consistency}
Please
generate yes/no questions that assess the physical plausibility and
consistency of object and scene motion.

\medskip
\textbf{Instruction:} You are an expert in physics and spatiotemporal
reasoning, with deep knowledge of real-world motion and dynamics in 4D
space (3D + time).

\medskip
\textbf{Note:}
\begin{enumerate}[leftmargin=*,nosep]
  \item Evaluation Goal: Assess the motion consistency of a 4D generation
        model, focusing on whether object and scene dynamics evolve
        plausibly over time.
  \item Core Evaluation Principles:
  \begin{enumerate}[leftmargin=*,nosep,label*=\arabic*.]
    \item Temporal Continuity: Are object trajectories and transformations
          smooth over time, without temporal flickering or abrupt
          discontinuities?
    \item Inter-object Interaction: Are interactions (e.g., collisions,
          pushes, pulls) physically reasonable and temporally aligned?
    \item Speed and Acceleration Coherence: Are velocity and acceleration
          patterns consistent with the object's mass, size, and environment?
    \item Scene-wide Consistency: Do all objects in the scene obey coherent
          motion logic, including global camera motion if present?
  \end{enumerate}
  \item Analyze the following video content and generate 5 specific yes/no
        questions that evaluate these motion rationality principles.

        \textit{Video content: \{content\}}
  \item Requirements:
  \begin{enumerate}[leftmargin=*,nosep,label*=\arabic*.]
    \item Generate exactly 5 questions covering the above principles.
    \item Each question should be answerable with yes/no.
    \item Questions must assess physical realism and motion logic.
    \item Focus on detecting unrealistic physics violations.
    \item Questions should be specific to the video content.
    \item The output should be a JSON list of strings.
  \end{enumerate}
\end{enumerate}
\end{promptbox}

\paragraph{Style Consistency.}
We measure style consistency using a VGG-based perceptual metric~\cite{duan2025worldscore}. For each clip $c$, we compute the Gram-matrix distance between the first and last frame of the clip:
\begin{equation}
e_{\text{style}}^{(c)} =
\left|
G(\mathbf{I}^{(c)}{1}) -
G(\mathbf{I}^{(c)}{T_c})
\right|_F,
\end{equation}
where $G(\cdot)$ denotes the Gram matrix of deep features.
The final score averages clip-level results:
\begin{equation}
e_{\text{style}} = 1- normalize(
\frac{1}{|\mathcal{C}|}
\sum_{c \in \mathcal{C}}
e_{\text{style}}^{(c)}).
\end{equation}

\begin{figure*}[htbp]
  \centering
   \includegraphics[width=1\linewidth]{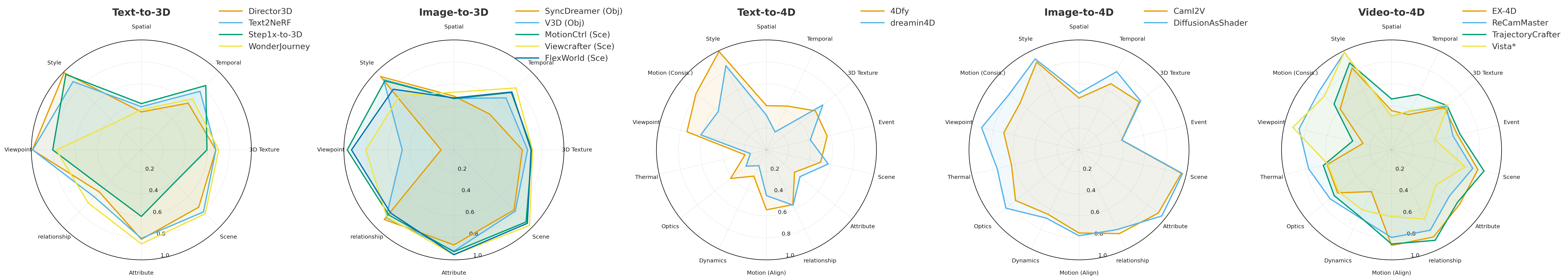}
   \vspace{-3mm} 
   \caption{Performance comparison for multiple world generative models.}
   \vspace{-3mm} 
   \label{fig:con_text}
\end{figure*}

\section{More Details for Experiment Metrics}

\textbf{SRCC and PLCC.}
We quantify the agreement between our benchmark metric and human subjective scores using the Pearson linear correlation coefficient (PLCC) and the Spearman rank-order correlation coefficient (SRCC). PLCC measures the linear correlation between the predicted scores and human ratings, while SRCC evaluates their monotonic relationship. Higher PLCC and SRCC values (closer to 1) indicate that the metric is more consistent with human perception.

\section{More Analysis for 3D/4D Generation Models}

\paragraph{Camera-control evaluation.}
As shown in Tab.~\ref{tab:camera_control}, our method \textit{ReCamMaster} achieves the highest normalized camera-control score. For \textit{TrajCrafter}, we observe that horizontal motions (pan left / pan right) are often entangled with unintended rotational changes of the viewing direction, leading to noticeable drift from the target trajectory and thus a lower camera-control score. In contrast, \textit{FlexWorld} tends to follow the target path reasonably well at the beginning of the sequence, but exhibits off-trajectory rotations in the later part of the video, which also degrades its overall performance. 

For all methods, we generate camera trajectories using a set of six canonical motion primitives: \textit{up}, \textit{down}, \textit{pan left}, \textit{pan right}, \textit{zoom in}, and \textit{zoom out}. The superior performance of \textit{ReCamMaster} indicates a better ability to preserve these intended motions without introducing spurious rotations or deviations from the desired camera path.

\begin{table}[h]
    \centering
    \begin{tabular}{l c}
        \hline
        \textbf{Method} & \textbf{Score} \\
        \hline
        CamI2V~\cite{zheng2024cami2v}          & 0.834 \\
        DiffusionAsShader~\cite{gu2025diffusion} & 0.793 \\
        FlexWorld~\cite{chen2025flexworld}
                                            & 0.528 \\
        MotionCtrl~\cite{wang2024motionctrl}                              & 0.798 \\
        ReCamMaster~\cite{bai2025recammaster}                     & 0.889 \\
        TrajCrafter~\cite{yu2025trajectorycrafter}                             & 0.700 \\
        WonderJourney~\cite{yu2024wonderjourney}                           & 0.825 \\
        \hline
    \end{tabular}
    \caption{Camera control scores (higher is better).}
    \label{tab:camera_control}
\end{table}

\paragraph{Multi-dimensional comparison of 3D/4D generators.}
Across the five radar plots in Fig.~\ref{fig:con_text}, we observe several
consistent strengths and weaknesses shared by current 3D and 4D world
generative models. On the positive side, most methods already achieve
relatively strong scores on \emph{style consistency}, \emph{scene control} and
\emph{attribute control}: both text- and image-conditioned 3D models form
near-saturated contours on these axes, and the best 4D models largely preserve
object identity and global scene composition when moving from static frames to
video. Viewpoint consisrency is also reasonably well handled for 3D generation and
video-to-4D, suggesting that multi-view geometry priors are effectively
exploited by many recent approaches.

In contrast, several dimensions remain challenging across modalities and
conditioning types.  Motion-related metrics—including \emph{motion
alignment}, \emph{dynamics} and \emph{motion consistency}—show a clear gap
between the best-performing models and the ideal score, revealing difficulties
in generating trajectories that are both semantically correct and physically
plausible. For physical realism, such as
\emph{optics} and \emph{thermal} cues, remain moderate for
4D generation, suggesting that current
architectures struggle to understanding physical law under complex
viewpoint and illumination changes. Finally, \emph{event control} and
\emph{relationship control} understanding are weak for text-to-4D,
highlighting an open problem of grounding rich textual descriptions into
coherent interactions over time. Overall, while style and
scene-level semantics are largely conquered, temporal reasoning, motion
control, and physically realism emerge as common directions for
improving future 3D/4D world models.

\begin{table}[t]
  \centering
  \caption{Correlation results on our improved 4D-condition Alignment}
  \resizebox{\linewidth}{!}{
  \begin{tabular}{lcc}
    \hline
    Method & Attribute Control PLCC &  Attribute Control SRCC  \\
    \hline
    
    Baseline & 0.167 & 0.236  \\
    Ours     & 0.483 & 0.443  \\
    \hline
  \end{tabular}}
  \label{tab:94videos}
\end{table}

\begin{table}[t]
  \centering
  \caption{Correlation results on our improved 4D consistency.}
  \resizebox{\linewidth}{!}{
  \begin{tabular}{lcc}
    \hline
    Method  & Style Consistency PLCC & Style Consistency SRCC \\
    \hline
    
    Baseline  & 0.383 & 0.457 \\
    Ours      & 0.545 & 0.457 \\
    \hline
  \end{tabular}}
  \label{tab:94videos1}
\end{table}

\section{More ablations for benchmark metrics}
We conduct a user study with 10 participants who provide subjective ratings on attribute control and style consistency for 100 generated videos. These ratings are averaged as the ground-turth score for our metric selection.

As summarized in Tab.~\ref{tab:94videos}, we compare our redesigned metric against a \emph{Baseline} setting, which use llava-video~\cite{zhang2024video,vbench2025} for question answering. And in Tab.~\ref{tab:94videos1}, we compare our redesigned metric against a \emph{Baseline} setting, which compute consistency metric at the video-level~\cite{duan2025worldscore}.

\paragraph{4D-Condition Alignment.}
Tab.~\ref{tab:94videos} demonstrates that our redesigned metric achieves significantly higher correlation with human judgment compared to the previous configuration. For attribute alignment, our method achieves a PLCC of 0.483 and SRCC of 0.443, representing a substantial gain over the previous setting (PLCC: 0.167, SRCC: 0.236). This improvement indicates that our enhanced question–answer formulation and the use of Keye-VL lead to more accurate semantic understanding, resulting in metrics that better reflect human perception.
\paragraph{Style Consistency.}
On style consistency, our clip-based evaluation approach further improves correlation with human assessments. Specifically, we observe a PLCC increase from 0.383 to 0.545, while the SRCC remains stable (Previous: 0.457, Ours: 0.457). The improved PLCC suggests that our modified video clip-based evaluation yields more stable and reliable style predictions, especially under varying motion and frame rate conditions, despite SRCC saturation. This highlights the benefit of localized, temporal-aware scoring in capturing perceptual appearance stability across generated videos.


\end{document}